\documentclass[conference]{IEEEtran}

\usepackage{graphicx}
\usepackage{booktabs}
\usepackage{amsmath}
\usepackage{float}
\usepackage{textcomp}
\usepackage[table]{xcolor}
\usepackage{multirow}
\usepackage{diagbox}
\usepackage{soul}
\usepackage{caption}
\usepackage[font=footnotesize]{caption}
\usepackage{orcidlink}
\usepackage{hyperref}
\usepackage{wrapfig}

\usepackage{colortbl}

\title{Surformer v1: Transformer-Based Surface Classification Using Tactile and Vision Features}

\author{
    \IEEEauthorblockN{
        Manish Kansana\IEEEauthorrefmark{1},
        Elias Hossain\IEEEauthorrefmark{1}\orcidlink{0000-0002-9364-7916},
        Shahram Rahimi\IEEEauthorrefmark{2},
        Noorbakhsh Amiri Golilarz\IEEEauthorrefmark{2}
    }
    \IEEEauthorblockA{\IEEEauthorrefmark{1}Department of Computer Science and Engineering, Mississippi State University, MS, USA\\
    \{mk1684, mh3511\}@msstate.edu}
    \IEEEauthorblockA{\IEEEauthorrefmark{2}Department of Computer Science, The University of Alabama, Tuscaloosa, AL, USA\\
    \{srahimi1, noor.amiri\}@ua.edu}
}

\begin{document}

\maketitle

\begin{abstract}
Surface material recognition is a key component in robotic perception and physical interaction, particularly when leveraging both tactile and visual sensory inputs. In this work, we propose Surformer v1, a transformer-based architecture designed for surface classification using structured tactile features and Principal Component Analysis (PCA)-reduced visual embeddings extracted via ResNet 50. The model integrates modality-specific encoders with cross-modal attention layers, enabling rich interactions between vision and touch. Currently, state-of-the-art deep learning models for vision tasks have achieved remarkable performance. With this in mind, our first set of experiments focused exclusively on tactile-only surface classification. Using feature engineering, we trained and evaluated multiple machine learning models, assessing their accuracy and inference time. We then implemented an encoder-only Transformer model tailored for tactile features. This model not only achieves the highest accuracy, but also demonstrated significantly faster inference time compared to other evaluated models, highlighting its potential for real-time applications. To extend this investigation, we introduced a multimodal fusion setup by combining vision and tactile inputs. We trained both Surformer v1 (using structured features) and Multimodal CNN (using raw images) to examine the impact of feature-based versus image-based multimodal learning on classification accuracy and computational efficiency. The results showed that Surformer v1 achieved 99.4\% accuracy with an inference time of 0.77 ms, while the Multimodal CNN achieved slightly higher accuracy but required significantly more inference time. These findings suggest that Surformer v1 offers a compelling balance between accuracy, efficiency, and computational cost for surface material recognition. The results also underscore the effectiveness of integrating feature learning, cross-modal attention and transformer-based fusion in capturing the complementary strengths of tactile and visual modalities.
\end{abstract}


\section{Introduction}
 Robots interacting with the physical world must be able to accurately perceive and classify the surfaces they touch. This capability is vital for safe object manipulation, navigation over varied terrains, and executing fine-grained tasks in unstructured environments. Tactile sensing provides detailed information about surface compliance, friction, and texture, properties that vision alone may struggle to capture reliably, especially under occlusions, poor lighting, or specular reflections. On the other hand, vision offers global context and appearance cues, making the combination of visual and tactile sensing a powerful solution for material recognition and robotic perception.

%Recent research has explored both vision-only models for image recognition \cite{simonyan2014very,dosovitskiy2020image} and tactile-only models for robust grasping \cite{li2020review,hu2024learning}, as well as various attempts at multimodal fusion \cite{yang2022touch,zhang2024multimodal,xue2023dynamic}. Among these, Convolutional Neural Networks (CNNs) have been widely used for end-to-end image-based classification, while graph attention and spiking neural networks have been applied to tactile signals \cite{chen2024tactile,yang2021robot}. More recently, transformer-based models have shown great promise in both unimodal and multimodal learning, particularly in vision-language applications \cite{liu2021swin,lu2019vilbert}, suggesting their potential for cross-sensory integration in robotics. Models such as MAE \cite{he2022masked} and CLIP \cite{radford2021learning} demonstrate the effectiveness of attention-based learning in fusing distributed representations from different input spaces.
Recent research has extensively explored both vision-based and tactile-based models. Vision-only models \cite{simonyan2014very, dosovitskiy2020image}, particularly those leveraging deep Convolutional Neural Networks (CNNs) and vision transformers, have achieved strong performance on image recognition tasks. However, their reliance on visual features makes them sensitive to occlusion, illumination variations, and specular noise. On the other hand, tactile-only models have demonstrated robustness in manipulation, object understanding, grasping and tactile data classification tasks where physical contact is critical \cite{li2020review, hu2024learning, chen2024tactile, yang2021robot}. For example, GelSight-based tactile sensors have enabled the extraction of high-resolution surface deformations, which are highly informative for material property inference.

To harness the strengths of both sensing modalities, multimodal learning strategies have been introduced to improve surface and object understanding. Early efforts employed simple feature-level fusion (e.g., concatenation) or decision-level fusion techniques \cite{atrey2010multimodal, srivastava2012multimodal}, but these often failed to capture the nuanced relationships between heterogeneous modalities. More advanced fusion architectures have been proposed, including those using attention-based networks and cross-modal deep representations \cite{nagrani2021attention, lee2022cross, xu2017learning}. Models such as \cite{calandra2018more, miller1999integration, struckmeier2019vita} have been developed for grasp planning, object exploration, and even navigation while interacting with aliased environments highlighting the complementary nature of tactile and visual data.
In parallel, transformer-based models have emerged as a powerful backbone for multimodal representation learning due to their ability to model long-range dependencies and perform cross-modal reasoning. Architectures such as \cite{liu2021swin, lu2019vilbert} have shown remarkable generalization in vision-language tasks, suggesting their potential for cross-sensory integration in robotics. Moreover, models like MAE \cite{he2022masked} and CLIP \cite{radford2021learning} demonstrate the effectiveness of attention-based learning in fusing distributed representations from different input spaces.

%Despite these advances, several limitations persist. First, end-to-end deep models often require large datasets and are difficult to interpret, particularly in tactile domains where data is less abundant. Second, simple fusion strategies, such as feature concatenation or prediction averaging, fail to capture the complex relationships between modalities. Third, few models offer a transparent and modular way to incorporate both structured tactile information and learned visual embeddings. 

 Despite these advances, several critical limitations persist in current surface classification approaches. First, end-to-end deep learning models typically rely on large volumes of labeled data, which are often scarce in tactile domains due to the complexity and cost of sensor setup and manual annotation. This data dependency not only limits scalability but also hampers generalizability in real-world scenarios. Second, many existing multimodal methods employ simplistic fusion strategies, such as feature concatenation or prediction averaging, which are inadequate for modeling the rich, nonlinear relationships and contextual dependencies between tactile and visual modalities. These approaches often ignore the dynamic interactions that naturally occur when humans perceive and interpret materials through both vision and touch. Third, most prior models lack modularity and transparency to incorporate both structured tactile information and learned visual embeddings and often treat tactile and visual inputs as homogeneous data streams without accounting for their distinct structures and complementary roles. As a result, there remains a need for models that are both effective and computationally efficient, particularly for resource-constrained robotic platforms.

To address these limitations, we propose \textbf{Surformer v1}, a transformer-based architecture designed specifically for surface classification. The model processes structured tactile features alongside reduced visual embeddings, combining them through a unified mid-level fusion framework. Unlike earlier methods, Surformer v1 enables cross-modal representation learning while maintaining computational efficiency, ensuring scalability to real-world robotic environments.
Our key contributions are as follows:

\begin{itemize}
    \item We introduce a multimodal architecture, leveraging both tactile and PCA-reduced ResNet 50-based visual embeddings suitable for robotic deployment and extension to other sensor combinations.
    \item We employ a mid-level fusion architecture with multi-head cross-attention layers to learn complementary representations, joint embeddings, and bidirectional interaction between tactile and visual inputs.
    \item We implement a lightweight,  encoder only Transformer for tactile-only classification, achieving the highest accuracy and the lowest latency, making it suitable for real-time deployment on embedded robotic systems where computational resources are limited.
    \item We evaluate our models on the \textbf{Touch and Go} dataset \cite{yang2022touchandgo} and demonstrate that multimodal fusion improves accuracy over tactile-only baselines.
    %\item We provide a modular and lightweight framework suitable for real-time robotic deployment and extension to other sensor combinations. (I'm not sure if its lightweight)
\end{itemize}

The rest of this paper is organized as follows: Section II details the Surformer v1 architecture. Section III describes the dataset and preprocessing pipeline, experimental results and baseline comparisons. Finally, Section IV concludes the paper and discusses future work.

% ===========================================
% FIGURE 1: Top 7 tactile feature importance
% ==========================================

\begin{figure}[!htbp]
    \centering
    \includegraphics[width=0.98\linewidth]{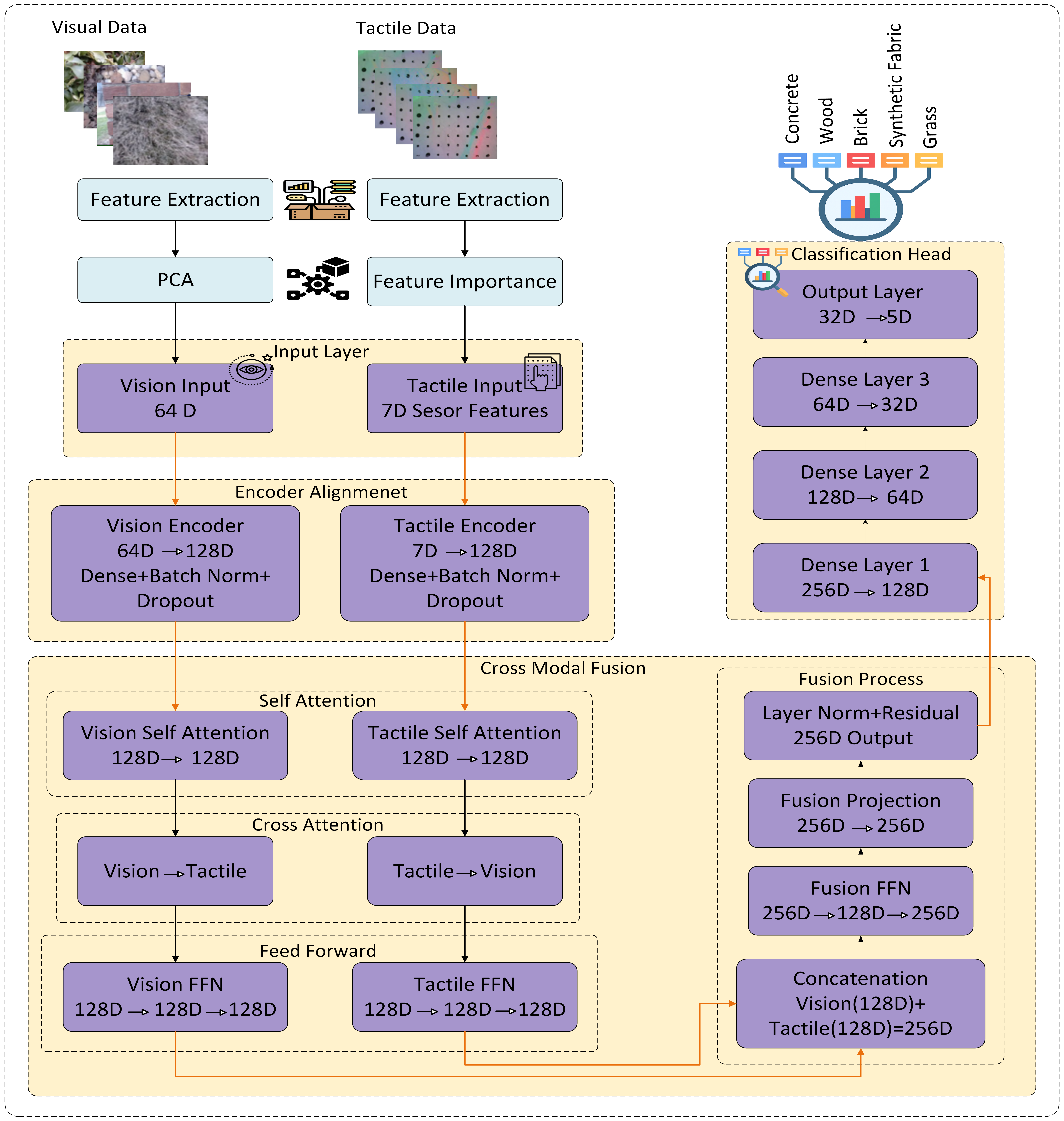}
    \caption{Architecture of Surformer v1 for multi-modal surface classification from tactile and visual data. The model accepts two input modalities: (1) visual features extracted from surface images using a pretrained ResNet-50 model and reduced to 64 dimensions via PCA, and (2) the top 7 tactile features selected based on importance analysis. Each input passes through a modality-specific encoder to map it into a unified 128-dimensional latent space using dense layers, batch normalization, and dropout. In the Cross-Modal Fusion stage, both self-attention and bidirectional cross-attention mechanisms are applied to capture intra- and inter-modality relationships. Feed-forward networks refine features within each modality before they are concatenated and passed through a fusion block involving residual connections, projection layers, and normalization. The fused representation is finally processed by a classification head comprising three fully connected layers and a softmax output layer to predict one of five surface material classes. This design enables the model to leverage complementary cues from both modalities, improving surface understanding beyond single-sensor methods.}
    \label{fig:fig_surformer_architecture}
    \vspace{-0.3cm}
\end{figure}

% ===========================================
% FIGURE 2: Bar Chart of Feature Importance
% ==========================================
% \begin{figure*}[t]
% \centering
% \includegraphics[width=0.7\textwidth]{fig2.png}
% \caption{Top 7 tactile feature importance scores.}
% \label{fig:top-tactile-feature}
% \end{figure*}
\begin{figure}[!htbp]
\begin{minipage}[t]{0.48\textwidth}
    \includegraphics[width=\textwidth]{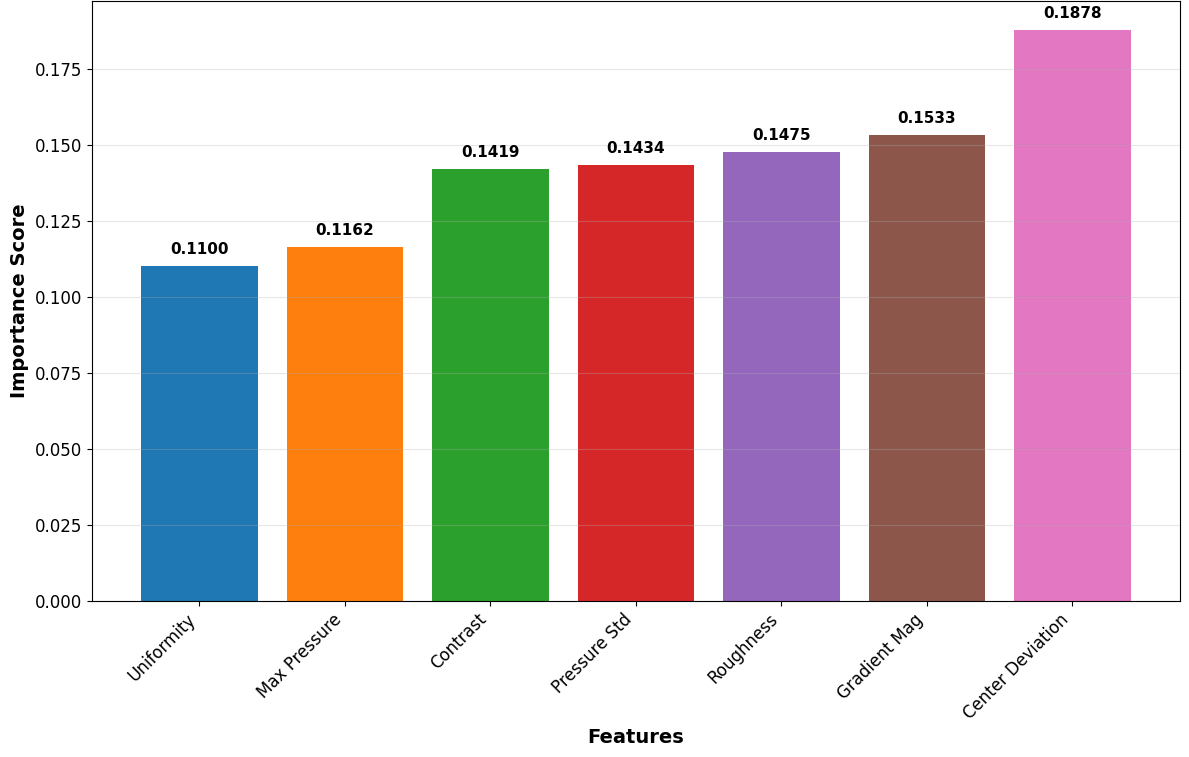}
    \caption{Top 7 tactile feature importance scores.}
    \label{fig:top-tactile-feature}
\end{minipage}
\end{figure}

\section{Methods}
%For tactile-only classification, four classical machine learning models and one deep learning model were evaluated for surface/material classification. The machine learning models included Random Forest, XGBoost, and Support Vector Machines (with both RBF and linear kernels). For deep learning, a simple Transformer model was used. To evaluate and rank feature importance, only the Random Forest model was used.

 In this section we discuss Surformer v1 architecture, which is designed for surface material classification, leveraging modality-specific encoders and cross-modal attention to jointly learn discriminative representations from tactile and visual inputs. Tactile inputs consist of structured, low-dimensional features extracted from touch and go dataset which was originally collected using GelSight sensors \cite{johnson2009retrographic}, while visual features are derived from ResNet50 \cite{he2016deep} and reduced via PCA to maintain balance between modalities. As illustrated in Fig. \ref{fig:fig_surformer_architecture}, the overall model architecture consists of four key stages: (1) feature processing, (2) modality specific encoders, (3) cross-modal fusion blocks that enable bidirectional information flow and fusion between modalities, and (4) a classification head that outputs surface material probabilities.

\subsection{Feature Processing}
\subsubsection{Feature Extraction}

For tactile features, GelSight sensor images were first converted to grayscale and processed through a custom feature engineering pipeline designed to extract key characteristics. Specifically, two distinct sets of features were derived: \textbf{Texture/Roughness Features}, which capture surface-level texture properties, and \textbf{Pressure/Contact Features}, which characterize the contact dynamics during tactile interaction.
The Texture/Roughness Features include surface roughness (deviation), gradient magnitude, local contrast, edge density, and uniformity index. The Pressure/Contact Features consist of average pressure, maximum pressure, contact area, pressure standard deviation, and center of pressure deviation. These features were extracted to capture important information that are essential for better classification. A detailed description of these features is provided in Table \ref{tab:tactile_features}. For the vision modality, raw RGB images are first preprocessed by scaling the pixel values to the [0,255] range and resizing them to 224×224×3 to match the input requirements for ResNet50. The images are then passed through a pre-trained ResNet50 backbone (excluding the top classification layers), originally trained on ImageNet, followed by Global Average Pooling to convert the 7×7×2048 feature maps into 2048-dimensional dense embeddings that capture rich visual representations including textures, colors, and structural patterns crucial for material identification.

\begin{table*}[!t]
\centering
\captionsetup{font=footnotesize}
\caption{Description of tactile features used for surface classification.}
\vspace{-0.2cm}

\begin{tabular}{p{4.5cm} p{12.0cm}}  % Adjusted column widths for full span
\toprule
\rowcolor{green!20}
\textbf{Texture/Roughness Feature} & \textbf{Description} \\
\midrule
Gradient Magnitude & Sharpness or local variations in surface texture; helps distinguish coarse vs. smooth textures \\
Contrast & Variation in intensity across the tactile surface; differentiates flat from textured materials \\
Roughness & Surface unevenness; distinguishes materials like concrete (high) vs. synthetic fabric (moderate) \\
Uniformity & How evenly the tactile pressure is distributed; identifies consistent contact textures \\
Edge Density & The number of distinct edges or transitions per unit area \\
\midrule
\rowcolor{green!20}
\textbf{Pressure Feature} & \textbf{Description} \\
\midrule
Pressure Std & Variability in pressure over the contact area; surface irregularity or material compliance \\
Center Deviation & How far the pressure centroid deviates from geometric center; reflects asymmetric or uneven contacts \\
Max Pressure & Maximum localized pressure during contact; associated with material hardness or contact sharpness \\
Avg Pressure & Overall pressure magnitude across contact area; related to surface hardness and grip force \\
Contact Area & Total area activated during contact; depends on material softness and indentation profile \\
\bottomrule
\end{tabular}
\label{tab:tactile_features}
\end{table*}

\begin{table*}[t]
\centering
\captionsetup{font=footnotesize}
\caption{Feature importance rankings for texture/roughness and pressure/contact sets using Random Forest.}
\vspace{-0.2cm}
\footnotesize
\begin{tabular}{p{4.6cm} p{1.5cm} p{4.6cm} p{1.5cm}}
\toprule
\rowcolor{green!20}
\textbf{Texture/Roughness Feature} & \textbf{Importance} & \textbf{Pressure/Contact Feature} & \textbf{Importance} \\
\midrule
Gradient Magnitude & 0.2493 & Pressure Std        & 0.3390 \\
Contrast           & 0.2262 & Center Deviation    & 0.2001 \\
Roughness          & 0.2244 & Max Pressure        & 0.1942 \\
Uniformity         & 0.1733 & Avg Pressure        & 0.1745 \\
Edge Density       & 0.1268 & Contact Area        & 0.0922 \\
\bottomrule
\end{tabular}
\label{tab:combined_importance}
\end{table*}

\subsubsection{Feature Selection}
For tactile features, both Texture/Roughness Features and Pressure/Contact Features are evaluated and ranked with Random Forest (see Table \ref{tab:combined_importance}) to gain insight on the features and their importance. The top seven features were selected based on their importance scores, and evaluated with RF again as illustrated in Fig. \ref{fig:top-tactile-feature}, and concatenated into a unified 1D feature vector. This vector served as input for downstream surface classification tasks.

\begin{figure}[H]
    \centering
    \vspace{0.5cm}
    \includegraphics[width=0.99\linewidth]{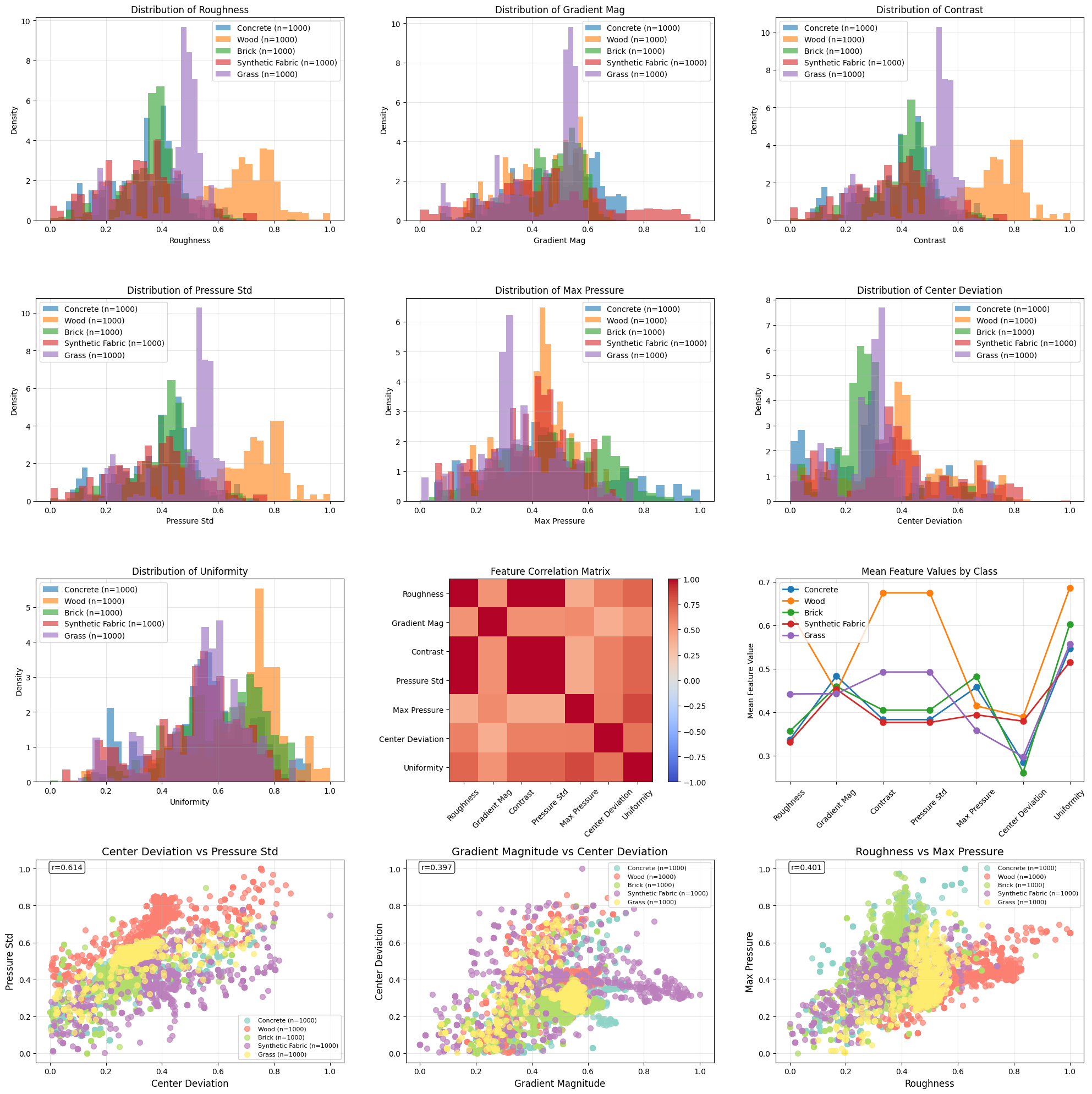}
    \caption{Tactile Features Analysis. The figure shows the distribution of each feature's density across all five classes and includes scatter plots to visualize relationships between pairs of features. These plots help reveal class separability and feature interactions.}
    \label{fig:fig3_tactile_features}
    \vspace{-0.3cm}  % Slight adjustment if needed
\end{figure}

To better understand the discriminative capacity of the selected tactile features and how they capture different physical properties relevant to material classification, we performed a detailed statistical analysis of the top seven features identified via Random Forest-based feature importance ranking. These features include Roughness, Gradient Magnitude, Contrast, Pressure Standard Deviation, Maximum Pressure, Center Deviation, and Uniformity. Fig. \ref{fig:fig3_tactile_features} presents the feature distributions across all five surface material classes (Concrete, Wood, Brick, Synthetic Fabric, and Grass), highlighting class-specific trends. A pairwise correlations reveals that most features are weakly correlated, indicating that each captures distinct surface properties. The mean feature profile per class confirms that the selected features carry discriminative patterns essential for material classification, thereby justifying their use as structured input for the Surformer v1 model. Additionally, pairwise scatter plots illustrate useful feature interactions, such as the relationship between roughness and maximum pressure or between gradient magnitude and center deviation, which further support the separability of classes in the feature space. 

For the vision features, PCA was applied to reduce the dimensionality from 2048 to a more compact representation with 64 dimensions. The reduced embeddings retained  90.7 \% variance. As a result, the final visual embeddings used for downstream tasks had a shape of (5000, 64).

%\subsection{Tactile Only Classification}
%To classify the surface material based on the extracted features, we adopt a transformer encoder architecture for 1D tactile features vectors. The model begins with an input layer that maps the features to a 64 latent dimension using a linear transformation. Positional information is encoded via learnable positional embeddings to preserve order dependencies across the feature vectors. The Transformer encoder consist of 3 layers and 8 attention heads. For classification, the encoder output passes through a head comprising Layer layer, dropout, a linear transformation from 64 to 32 dimensions, a ReLU activation, and a final linear layer mapping to the 5 material classes. \Amiri{I moved this to result section} 

% =====================================
% FIGURE 3: Tactile feature analysis
% =====================================

\subsection{Multi-modal Classification}

\subsubsection{Cross-modal Fusion}
Cross-modal fusion block is the core of Surformer v1 architecture, exchanging information between vision and tactile features. Each fusion block performs multi-stage attention-based integration, allowing the model to classify the material. Before the fusion block, each input modality is passed through a dedicated encoder comprising progressive dense layers. These act as input embedding layers, mapping tactile features (7D) and reduced visual features (64D using PCA) into a common 128-dimensional latent space. This shared embedding space enables an effective interaction in subsequent attention-based fusion. 

\paragraph{\textbf{Self Attention and Cross Attention:}}
The block begins by processing each input independently through self-attention mechanisms, where vision features attend to other vision features and tactile features attend to other tactile features. This self-attention stage allows each modality to refine its internal representations by identifying the most relevant patterns within itself. Following self-attention, the cross-attention mechanism operates bidirectionally: vision features serve as queries that attend to tactile features as keys and values, while simultaneously tactile features serve as queries that attend to vision features. The attention mechanisms within the fusion block leverage multi-head attention with 2 heads, each operating on 64-dimensional subspaces of the 128-dimensional feature space. This multi-head design is crucial because it allows the model to capture different types of cross-modal relationships simultaneously. The diversity of attention heads ensures comprehensive cross-modal understanding. The mathematical foundation of the cross-modal attention follows the scaled dot-product attention formula, but is adapted for cross-modal interaction. When vision features query tactile information, the attention weights are computed as:

\begin{equation}
\text{Attention}(\mathbf{Q}_{\text{v}}, \mathbf{K}_{\text{t}}, \mathbf{V}_{\text{t}}) = \text{softmax}\left( \frac{\mathbf{Q}_{\text{v}} \mathbf{K}_{\text{t}}^\top}{\sqrt{d_k}} \right) \mathbf{V}_{\text{t}}
\end{equation}

where, \(\mathbf{Q}_{\text{v}}\) refers to the query matrix derived from the visual modality, \(\mathbf{K}_{\text{t}}\) denotes the key matrix from the tactile modality, and \(\mathbf{V}_{\text{t}}\) is the value matrix also from the tactile modality. The dot product \(\mathbf{Q}_{\text{v}} \mathbf{K}_{\text{t}}^\top\) computes the attention scores, scaled by \(\sqrt{d_k}\), the dimensionality of the keys. The softmax function transforms these scores into attention weights, which are used to compute a weighted sum over the tactile values, enabling cross-modal feature alignment.
This formulation allows vision features to selectively attend to relevant tactile patterns based on their current visual understanding, creating a dynamic information exchange that adapts to the specific characteristics of each input sample.

\paragraph{\textbf{Feed-Forward Processing and Normalization:}}

After the attention operations, fusion block applies feed-forward networks to both modalities independently. These FFNs process 128 dimensional features  through a 128-dimensional hidden layer back to 128 dimensions using ReLU activation functions. Layer normalization and residual connections are applied throughout the fusion block to ensure stable training and effective gradient flow. The normalization is applied before each sub-layer (pre-layer normalization). Residual connections allow the model to learn incremental refinements to the features while preserving the original information, preventing the vanishing gradient problem in deep networks.

\paragraph{\textbf{Fusion Process:}}

After the vision and tactile features have been refined through self-attention, cross-attention, and feed-forward processing, they are then concatenated to create a 256-dimensional combined representation. The concatenation preserves the distinct characteristics of each modality while creating a unified multi-modal feature vector that contains both visual and tactile information along with their correlations. The concatenated features then undergo a fusion feed-forward network that operates on the full 256-dimensional space (256D → 128D → 256D). This fusion FFN learns to integrate the combined multi-modal information. A projection layer then transforms these fused features, and the final output undergoes layer normalization with a residual connection from the original concatenated features, ensuring that the model can learn incremental refinements while preserving essential information.

\subsubsection{Classification Head}
A progressive dimensionality reduction strategy using multiple dense layers, each followed by batch normalization and dropout. The fused 256-dimensional multi-modal features are systematically reduced through a three-stage pathway: from 256 to 128, then to 64 dimensions, and finally to 32D. Each stage incorporates normalization and dropout to enhance training robustness. Finally, a softmax layer maps the 32-dimensional compressed representation to a probability distribution over the five material classes: Concrete, Wood, Brick, Synthetic Fabric, and Grass.

\section{Experimental Results and Discussion}

\subsection{Data Preprocessing}

The touch and go dataset \cite{yang2022touchandgo}, consist of synchronized vision and gel-sight sensor video for more than 20 labeled classes. Thousands of synchronized frames were extracted from both vision(video) and tactile (GelSight), with each frame labeled according to its corresponding surface class. These images were preprocessed by resizing them to 224 * 224 resolution with three RGB channels and normalizing pixel values between 0 and 1. For this study, a curated subset of the dataset without noise or incorrect labels was selected to ensure clean and reliable input data. The extracted classes included Concrete (288), Wood (364), Brick (661), Synthetic Fabric (364), and Grass (600). To address class imbalance, data augmentation was performed using Keras, resulting in a balanced final dataset of 5,000 paired vision and tactile images, 1,000 samples per class.
\\
 For the vision images, data augmentation was applied using a series of transformations, including a rotation range of 25 degrees, width and height shifts of up to 10\%, and a zoom range of 0.05. Horizontal flipping was enabled to increase variability, while vertical flipping was avoided to maintain the realism of natural scenes. Brightness variation was constrained within a conservative range of 0.8 to 1.2, and channel shifts were limited to 0.1. All transformations used the nearest fill mode, and no automatic rescaling was applied since preprocessing handled it separately.

For the tactile images, augmentation included a rotation range of 20 degrees, and width and height shifts up to 10\%. A zoom range of 0.05 was applied, with horizontal flipping enabled. Brightness adjustments were within a narrow range of 0.9 to 1.1. Similar to the vision image augmentation, no rescaling was performed, and the same \textquotedblleft{}nearest\textquotedblright{} fill mode was used.

\subsection{Training Strategy}

 For all the tactile only models including RF \cite{breiman2001random}, XGBoost\cite{chen2016xgboost}, SVM \cite{cortes1995support}, Transformer\cite{vaswani2017attention}, the dataset was split into 80\% training and 20\% testing using stratified sampling. For the Surformer v1 and Multi-modal CNN, dataset was split into 80\% training, 10\% testing, and 10\% validation to enable hyperparameter tuning. The encoder only transformer model was trained with a learning rate of 0.001 for 150 epochs using the AdamW optimizer with a batch size 64, and a learning rate scheduler (ReduceLROnPlateau) to adaptively reduce the learning rate based on validation performance with the patience of 10 and a factor of 0.5. Regularization techniques included a weight decay of 0.01 and dropout with a rate of 0.1 to prevent overfitting.

The multi-modal Surformer v1 was trained with a learning rate of 0.5e-5 for up-to 100 epochs using the Adam optimizer and a batch size of 32 to ensure stable convergence across both vision and tactile modalities. The EarlyStopping callback was used with a patience of 15 epochs based on validation accuracy, and a learning rate scheduler ReduceLROnPlateau was implemented to adaptively reduce the learning rate based on validation loss, with a patience of 8 epochs and a reduction factor of 0.5. The scheduler also included a minimum learning rate threshold of 1e-7. Regularization techniques included a dropout rate of 0.1 across all encoders, blocks, and classification heads. Batch normalization was employed in all dense layers within the modality specific encoders.

The Multimodal CNN uses a dual-stream architecture with two parallel ResNet50 backbones, vision and tactile. Both backbones are initialized with ImageNet weights and process 224×224×3 RGB images. After feature extraction, Global Average Pooling reduces spatial dimensions, followed by stream-specific dense layers with ReLU activation and dropout. The vision and tactile features are concatenated into a 512-dimensional fused representation, which is further processed through an additional dense layer before final classification into five material classes via a softmax layer. The model is optimized using Adam optimizer with a learning rate of 1e-6, trained for 100 epochs with a batch size of 32, matching the training settings used for Surformer v1. Regularization techniques include L2 weight decay of 1e-4, dropout of 0.5, early stopping with a patience of 15 epoch based on validation accuracy, and learning rate reduction on plateau with a factor 0.5 for every 8 epoch on validation loss. Most hyper-parameters for Multi-model CNN were kept similar as Surformer v1.

Additionally, classical machine learning algorithms were carefully optimized and evaluated. The Random Forest classifier was configured with 200 decision trees, a maximum depth of 15, and additional constraints, including a minimum samples split of 5 and a minimum samples leaf of 2. The XGBoost model, known for its gradient-boosted tree ensembles, was employed with 300 estimators, a maximum depth of 8, a learning rate of 0.1. To enhance generalization, sub-sampling and column sampling were both set to 0.8 for regularization. Finally, Support Vector Machines (SVMs) were used with two kernel variants. The RBF kernel was configured with C = 10 and gamma = \textquotedblleft{}scale\textquotedblright{} to capture non-linear decision boundaries, while the linear kernel used C = 1.0 to model high-dimensional linear separability. Both SVM configurations were enabled with probability estimation to support downstream interpretability and confidence analysis. These classical models served as strong baselines and provided valuable insights into the discriminative power of extracted tactile features.

 %80\% training and 20\% testing with stratification enabled. For the Surformer, dataset was split into 80\% training, 20\% testing, then further splits the training set into 80\% training and 20\% validation. This ensures both vision and tactile features are split synchronously while maintaining class distribution through stratification. The final distribution of the dataset for Surformer is 64\% Training, 20\% testing, and 16\% validation, and all models are evaluated on the test set for accuracy and other metrics.
\begin{table*}[t]
\centering
\captionsetup{font=footnotesize}
\caption{Classification performance comparison: Tactile (T) Only vs Vision and Tactile (V-T) models.}
\vspace{-0.2cm}
\footnotesize
\begin{tabular}{clcccccc}
\toprule
& \textbf{Model} & \textbf{Precision} & \textbf{Recall} & \textbf{F1-Score} & \textbf{Accuracy} & \textbf{Parameters} & \textbf{Inference Time (ms)} \\
\midrule
\multirow{5}{*}{\rotatebox{90}{T-Only}} 
 & Random Forest             & 0.96 & 0.96 & 0.96 & 0.9560 & 108,788     & 0.2819 \\
 & XGBoost                   & 0.97 & 0.97 & 0.97 & 0.9670 & 1,500       & 0.0923 \\
 & SVM (RBF)                 & 0.87 & 0.87 & 0.86 & 0.8660 & N/A         & 0.0483 \\
 & SVM (Linear)              & 0.68 & 0.68 & 0.67 & 0.6810 & 80          & 0.0337 \\
 & Encoder-only Transformer & 0.97 & 0.97 & 0.97 & 0.9740 & 152,901     & 0.0085 \\
\midrule
\multirow{2}{*}{\rotatebox{90}{V-T}} 
 & Multimodal-CNN            & 1.00 & 1.00 & 1.00 & 1.0000 & 48,311,301  & 5.0737 \\
 & \textbf{Surformer v1}     & 0.99 & 0.99 & 0.99 & 0.9940 & 673,321     & 0.7271 \\
\bottomrule
\end{tabular}
\label{tab:combined_accuracy_comparison}
\end{table*}

\subsection{Results}
 In this section, we analyze both tactile-only and multi-modal classification approaches, comparing their performance based on precision, recall, F1-score, and accuracy. 

%For tactile only setup, the final feature vector, comprising extracted features from tactile  images was used as input for both machine and deep learning models. These models were then trained and evaluated on the test set to assess their ability to classify surface materials using tactile information exclusively.

%To classify the surface material based on the extracted features, we adopt an encoder only transformer architecture for 1D tactile features vectors. The model begins with an input layer that maps the features to a 64 latent dimension using a linear transformation. Positional information is encoded via learnable positional embeddings to preserve order dependencies across the feature vectors. This transformer consist of 3 layers and 8 attention heads. For classification, the encoder output passes through a head \Amiri{which head? what head?} comprising Layer layer \Amiri{what?? You are repeating the layer. Double check}, dropout, a linear transformation from 64 to 32 dimensions, a ReLU activation, and a final linear layer mapping to the 5 material classes. 

%As can be seen from Table 1, for tactile data the encoder-only transformer achieved the accuracy of 0.9740, followed closely by XGboost and Random Forest with accuracies of 0.9670, and 0.9560, respectively. While SVM with an RBF kernel performed moderately well with the accuracy of 0.8660, linear SVM yielded significantly lower results (accuracy of 0.6810), indicating its limited ability to model complex decision boundaries in tactile feature space. 
As shown in Table \ref{tab:combined_accuracy_comparison}, for tactile-only classification, the encoder-only Transformer achieved the highest accuracy of 0.9740, followed closely by XGBoost (0.9670) and Random Forest (0.9560). The SVM with an RBF kernel performed moderately well with an accuracy of 0.8660, while the linear SVM yielded significantly lower results (0.6810 accuracy), highlighting its limited ability to model the complex decision boundaries required for tactile feature classification. In addition to classification performance, inference time and model size were also evaluated. Inference time in this study refers to the time it takes for a trained model to process new, unseen data and produce a classification output. It was evaluated on the test set with a batch size of 100, and is reported in milliseconds per sample for both tactile-only and vision-tactile models. The encoder-only Transformer not only delivered strong accuracy but also demonstrated the lowest inference time (0.0085 ms per sample) among all models, making it highly suitable for real-time applications. In contrast, while Random Forest and XGBoost showed competitive accuracy, they exhibited higher inference times (0.2819 ms and 0.0923 ms, respectively) and, in the case of Random Forest, a substantially larger number of parameters (108,788 parameters).

For multimodal classification, Surformer v1 demonstrated strong performance, highlighting the effectiveness of cross-modal fusion for surface material recognition. It achieved scores of 0.99 in precision, recall, and F1-score, with an accuracy of 0.9940. We further implemented a Multimodal CNN using raw vision and tactile images and compared the results with Surformer v1. The Multimodal CNN achieved the highest accuracy (1.00) across all metrics but required substantially higher inference time (5.0737 ms) and had a much larger number of parameters (48.3 million) compared to Surformer v1, which maintained a faster inference time of 0.7271 ms with only 673,321 parameters. This result demonstrates that while the Multimodal CNN slightly outperforms in accuracy, Surformer v1 offers a significantly more efficient and compact solution, making it better suited for real-time robotic applications where computational resources are limited.

\section{Conclusion}

%In this paper, we introduced Surformer v1, a transformer-based model for surface material classification that leverages both tactile and visual modalities. By combining structured tactile features with PCA-reduced visual embeddings extracted from ResNet 50, Surformer v1 enables robust material recognition. 

%In this paper, we introduced Surformer v1, a transformer-based model that combines structured tactile features and PCA-reduced visual embeddings from ResNet-50 to achieve robust surface material classification. The model architecture incorporates modality-specific encoders and cross-modal attention mechanisms, facilitating rich interactions between sensory inputs and capturing complementary characteristics of each modality. Experimental results on the Touch and Go dataset demonstrate that our proposed multimodal approach significantly outperforms both conventional machine learning models (e.g., SVM, Random Forest, XGBoost) and unimodal transformer baseline. Notably, Surformer v1 (Tactile and Vision) achieved an accuracy of 99.7\%, validating the effectiveness of cross-modal attention in enhancing surface recognition tasks. 
%While Surformer v1 demonstrates promising results, several directions remain open for future research. We plan to evaluate the generalizability of Surformer v1 on other tactile sensor types and visual domains to test its adaptability. By enhancing the multimodal capabilities and scalability of Surformer v1, future iterations can further bridge the gap between human-like perception and robotic interaction.

 In this paper, we introduced Surformer v1, a transformer-based model that combines structured tactile features and PCA-reduced visual embeddings from ResNet-50 to achieve robust surface material classification. The model architecture incorporates modality-specific encoders and cross-modal attention mechanisms, facilitating rich interactions between sensory inputs and capturing complementary characteristics of each modality. We investigated surface material classification through both tactile-only and multimodal learning approaches. We began by focusing on tactile-only classification, where feature engineering enabled the training of several classical machine learning models. An encoder-only Transformer tailored for tactile features was also implemented, which, while achieving strong accuracy, demonstrated significantly faster inference times than other models, highlighting its suitability for real-time applications where efficiency is critical. To extend this investigation, we incorporated visual information and introduced a multimodal fusion framework combining both tactile and vision inputs. We implemented and compared two multimodal models: a Multimodal CNN trained on raw images and the proposed Surformer v1, which operates on structured feature representations. The results showed that while the Multimodal CNN achieved slightly higher classification accuracy, Surformer v1 delivered substantially faster inference time and a smaller number of parameters. These findings demonstrate that Surformer v1 provides a compelling trade-off between accuracy, computational efficiency, and real-time applicability. Overall, this study underscores the value of integrating structured feature learning with cross-modal attention for efficient and accurate surface material recognition. Future work will explore scaling the approach to larger and more diverse datasets, evaluating generalizability across sensor types, and further optimizing the architecture for deployment in real-world robotic systems.

% %%%%%%%%%%%%%%%%%%%%%%%%%%%%%%%%%%%%%%%%%%
% \authorcontributions{Conceptualization, M.K., E.H., N.A.G. and S.R.; methodology, M.K., and N.A.G; software, M.K. and E.H; validation, M.K., E.H., N.A.G. and S.R.; formal analysis, M.K; investigation, M.K., E.H., N.A.G. and S.R.; resources, M.K., N.A.G.; data curation, M.K.; writing---original draft preparation, M.K. and N.A.G.; writing---review and editing, N.A.G. and S.R.; visualization, M.K. and N.A.G; supervision, N.A.G. and S.R.; project administration, S.R.; funding acquisition, N.A.G. and S.R. All authors have read and agreed to the published version of the manuscript.}

\section*{Data Availability}

We used the publicly available Touch and Go dataset, which is accessible through the original publication \cite{yang2022touchandgo}.

% \section*{Acknowledgment}
% The authors acknowledge the support and resources provided by the Predictive Analytics and Technology Integration (PATENT) Laboratory at Mississippi State University and Intelligent Systems and Predictive Analytics (ISPA) Laboratory at The University of Alabama.

\section*{Conflict of Interest}
The authors declare no conflicts of interest.

\bibliographystyle{IEEEtran}
\bibliography{references}

\end{document}